\documentclass[10pt,twocolumn,letterpaper]{article}

\usepackage[accsupp]{axessibility} 
\usepackage[pagenumbers]{cvpr} 

\usepackage{graphicx}
\usepackage{amsmath}
\usepackage{amssymb}
\usepackage{booktabs}
\usepackage{bm}
\usepackage{multirow}
\usepackage{makecell}
\def\x{$\times$}
\usepackage{pifont}
\newcommand{\cmark}{\ding{51}}%
\newcommand{\xmark}{\ding{55}}%
\usepackage{xcolor}         
\definecolor{demphcolor}{RGB}{144,144,144}

\usepackage{colortbl}
\definecolor{mygray}{gray}{0.95}

\definecolor{baselinecolor}{gray}{.9}

\newlength\savewidth
\usepackage{fontawesome}

\usepackage{algorithm}
\usepackage{listings}
\DeclareFixedFont{\ttb}{T1}{txtt}{bx}{n}{6} 
\DeclareFixedFont{\ttm}{T1}{txtt}{m}{n}{6}  

\definecolor{codegreen}{rgb}{0,0.6,0}
\definecolor{codegray}{rgb}{0.5,0.5,0.5}
\definecolor{codepurple}{rgb}{0.58,0,0.82}
\definecolor{backcolour}{rgb}{0.95,0.95,0.92}

\lstdefinestyle{mystyle}{
    commentstyle=\color{codegreen},
    keywordstyle=\color{magenta},
    numberstyle=\tiny\color{codegray},
    stringstyle=\color{codepurple},
    basicstyle=\ttfamily\footnotesize,
    breakatwhitespace=false,         
    breaklines=true,                 
    captionpos=b,                    
    keepspaces=true,                 
    numbers=left,                    
    numbersep=5pt,                  
    showspaces=false,                
    showstringspaces=false,
    showtabs=false,                  
    tabsize=2
}

\usepackage[pagebackref,breaklinks,colorlinks]{hyperref}

\usepackage[capitalize]{cleveref}
\crefname{section}{Sec.}{Secs.}
\Crefname{section}{Section}{Sections}
\Crefname{table}{Table}{Tables}
\crefname{table}{Tab.}{Tabs.}

\def\cvprPaperID{8184} 
\def\confName{CVPR}
\def\confYear{2023}

\begin{document}

\title{Bidirectional Cross-Modal Knowledge Exploration for Video Recognition \\ with Pre-trained Vision-Language Models}


\author{%
Wenhao Wu$^{1,2}$\quad
Xiaohan Wang$^{3}$\quad
Haipeng Luo$^{4}$\quad
Jingdong Wang$^{2}$\quad
Yi Yang$^{3}$\quad
Wanli Ouyang$^{5,1}$\\
$^1$The University of Sydney \qquad $^2$Baidu Inc. \qquad $^3$Zhejiang University \\ 
$^4$University of Chinese Academy of Sciences  \qquad $^5$Shanghai AI Laboratory\\
{\tt\small whwu.ucas@gmail.com}
}

\maketitle

\begin{abstract}
Vision-language models (VLMs) pre-trained on large-scale image-text pairs have demonstrated impressive transferability on various visual tasks. Transferring knowledge from such powerful VLMs is a promising direction for building effective video recognition models.
However, current exploration in this field is still limited.
We believe that the greatest value of pre-trained VLMs lies in building a bridge between visual and textual domains.
In this paper, we propose a novel framework called \textbf{BIKE}, which utilizes the cross-modal bridge to explore bidirectional knowledge: i) We introduce the Video Attribute Association mechanism, which leverages the \texttt{Video-to-Text} knowledge to generate textual auxiliary attributes for complementing video recognition.
ii) We also present a Temporal Concept Spotting mechanism that uses the \texttt{Text-to-Video} expertise to capture temporal saliency in a parameter-free manner, leading to enhanced video representation.
Extensive studies on six popular video datasets, including Kinetics-400 \& 600, UCF-101, HMDB-51, ActivityNet and Charades, 
show that our method achieves state-of-the-art performance in various recognition scenarios, such as general, zero-shot, and few-shot video recognition. 
 Our best model achieves a state-of-the-art accuracy of 88.6\% on the challenging Kinetics-400 using the released CLIP model. The code is available at \url{https://github.com/whwu95/BIKE}.

\end{abstract}

\section{Introduction}
\label{sec:intro}

\begin{figure}
\begin{center}
\includegraphics[width=1\linewidth]{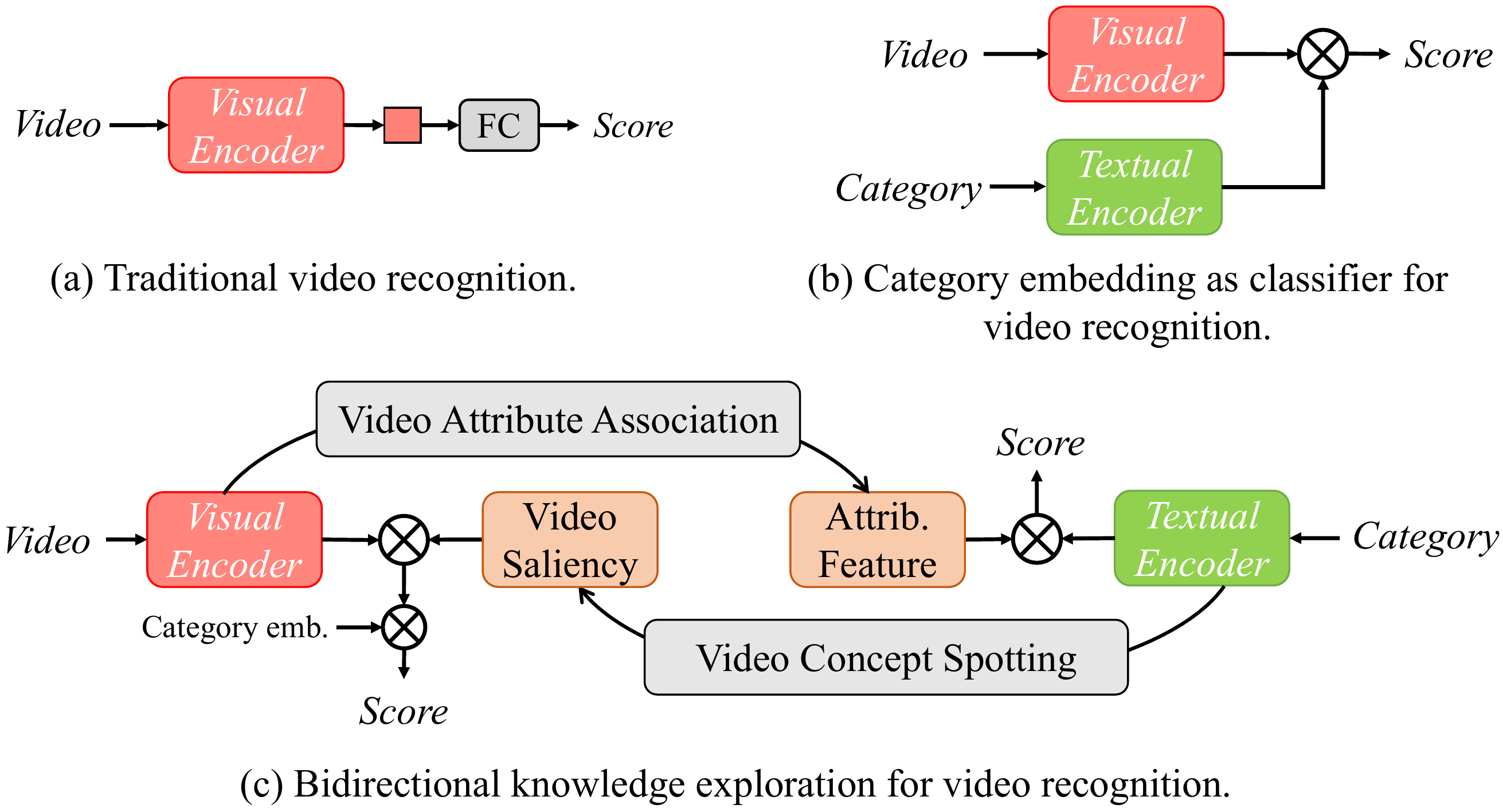}
\end{center}
\caption{Illustration of the difference between our paradigm (c) with existing unimodality paradigm (a) and cross-modal paradigm (b). Please zoom in for the best view.
}
\label{fig:overview}
\end{figure}



In recent years, the remarkable success of large-scale pre-training in NLP (\eg, BERT~\cite{devlin2018bert}, GPT~\cite{GPT,GPT3}, ERNIE~\cite{zhang2019ernie} and T5~\cite{2020t5}) has inspired the computer vision community. 
Vision-language models (VLMs) leverage large-scale noisy image-text pairs with weak correspondence for contrastive learning (\eg, CLIP \cite{CLIP}, ALIGN \cite{ALIGN}, CoCa \cite{yu2022coca}, Florence\cite{yuan2021florence}), and demonstrate impressive transferability across a wide range of visual tasks. 

Naturally, transferring knowledge from such powerful pre-trained VLMs is emerging as a promising paradigm for building video recognition models. 
Currently, exploration in this field can be divided into two lines. 
As depicted in Figure~\ref{fig:overview}(a), one approach~\cite{st-adaptor,evl,yang2023aim} follows the traditional unimodal video recognition paradigm, initializing the video encoder with the pre-trained visual encoder of VLM. Conversely, the other approach~\cite{wang2021actionclip,ju2022prompting,x-clip,text4vis} directly transfers the entire VLM into a video-text learning framework that utilizes natural language (\ie, class names) as supervision, as shown in Figure~\ref{fig:overview}(b).
This leads to an question: \emph{have we fully utilized the knowledge of VLMs for video recognition?}

In our opinion, the answer is No. The greatest charm of VLMs is their ability to build a bridge between the visual and textual domains. 
Despite this, previous research employing pre-aligned vision-text features of VLMs for video recognition has only utilized unidirectional video-to-text matching.
In this paper, we aim to facilitate bidirectional knowledge exploration through the cross-modal bridge for enhanced video recognition.
With this in mind, we mine \texttt{Video-to-Text} and \texttt{Text-to-Video} knowledge by 1) generating textual information from the input video and 2) utilizing category descriptions to extract valuable video-related signals.

In the first \texttt{Video-to-Text} direction, a common practice for mining VLM knowledge is to embed the input video and category description into a pre-aligned feature space, and then select the category that is closest to the video, as illustrated in Figure~\ref{fig:overview}(b), which serves as our baseline.
One further question naturally arises: \emph{Can we incorporate auxiliary textual information for video recognition?} To address this question, we introduce an \textbf{Video-Attributes Association} mechanism, which leverages the zero-shot capability of VLMs to retrieve the most relevant phrases from a pre-defined lexicon for the video. These phrases are considered potential ``attributes'' of the video and can predict the video category directly. For example, a video of someone kicking a soccer ball may be associated with relevant phrases such as ``running on the grass'', ``juggling soccer ball'' and ``shooting goal''. 
Surprisingly, using only the generated attributes, we can achieve 69\% top-1 accuracy on the challenging Kinetics-400 dataset.
Furthermore, these attributes provide additional information that the video visual signal may not capture, allowing us to build an \emph{Attributes Recognition Branch} for video recognition.

In the second \texttt{Text-to-Video} direction, we believe that temporal saliency in videos can be leveraged to improve video representations. For instance, in a video with the category ``kicking soccer ball'', certain frames of kicking the ball should have higher saliency, while other frames that are unrelated to the category or background frames should have lower saliency. 
This insight motivates us to propose the \textbf{Video Concept Spotting} mechanism, which utilizes the cross-model bridge to generate category-dependent temporal saliency. In previous works~\cite{wang2021actionclip,x-clip,text4vis}, this intuitive exploration was disregarded.
To be more specific, instead of treating each video frame equally, we use the correlation between each frame and the given concept (\eg, category) as a measure of frame-level saliency. This saliency is then used to temporally aggregate the frames, resulting in a compact video representation.

In the light of the above explorations, we propose \textbf{BIKE}, a simple yet effective framework via \textbf{BI}directional cross-modal \textbf{K}nowledge \textbf{E}xploration for enhanced video recognition. Our \textbf{BIKE} comprises two branches: the \emph{Attributes branch}, which utilizes the \textbf{Video-Attributes Association} mechanism to introduce auxiliary attributes for complementary video recognition, and the \emph{Video branch}, which uses the \textbf{Video Concept Spotting} mechanism to introduce temporal saliency to enhance video recognition.
To demonstrate the effectiveness of our \textbf{BIKE}, we conduct comprehensive experiments on popular video datasets, including Kinetics-400~\cite{kay2017kinetics} \& 600~\cite{k600}, UCF-101~\cite{ucf101}, HMDB-51~\cite{hmdb}, ActivityNet~\cite{caba2015activitynet} and Charades~\cite{charades}. The results show that our method achieves state-of-the-art performance in most scenarios, \eg, general, zero-shot, and few-shot recognition. 

Our main contributions can be summarized as follows:
\begin{itemize}
    \item We propose a novel framework called \textbf{BIKE} that explores bidirectional knowledge from pre-trained vision-language models for video recognition.
    \item In the \texttt{Video-to-Text} direction, we introduce the \textbf{Video-Attributes Association} mechanism to generate extra attributes for complementary video recognition.
    \item In the \texttt{Text-to-Video} direction, we introduce the \textbf{Video Concept Spotting} mechanism to generate temporal saliency, which is used to yield the compact video representation for enhanced video recognition.
\end{itemize}

\section{Methodology}

\begin{figure*}[t]
\begin{center}
\includegraphics[width=0.98\linewidth]{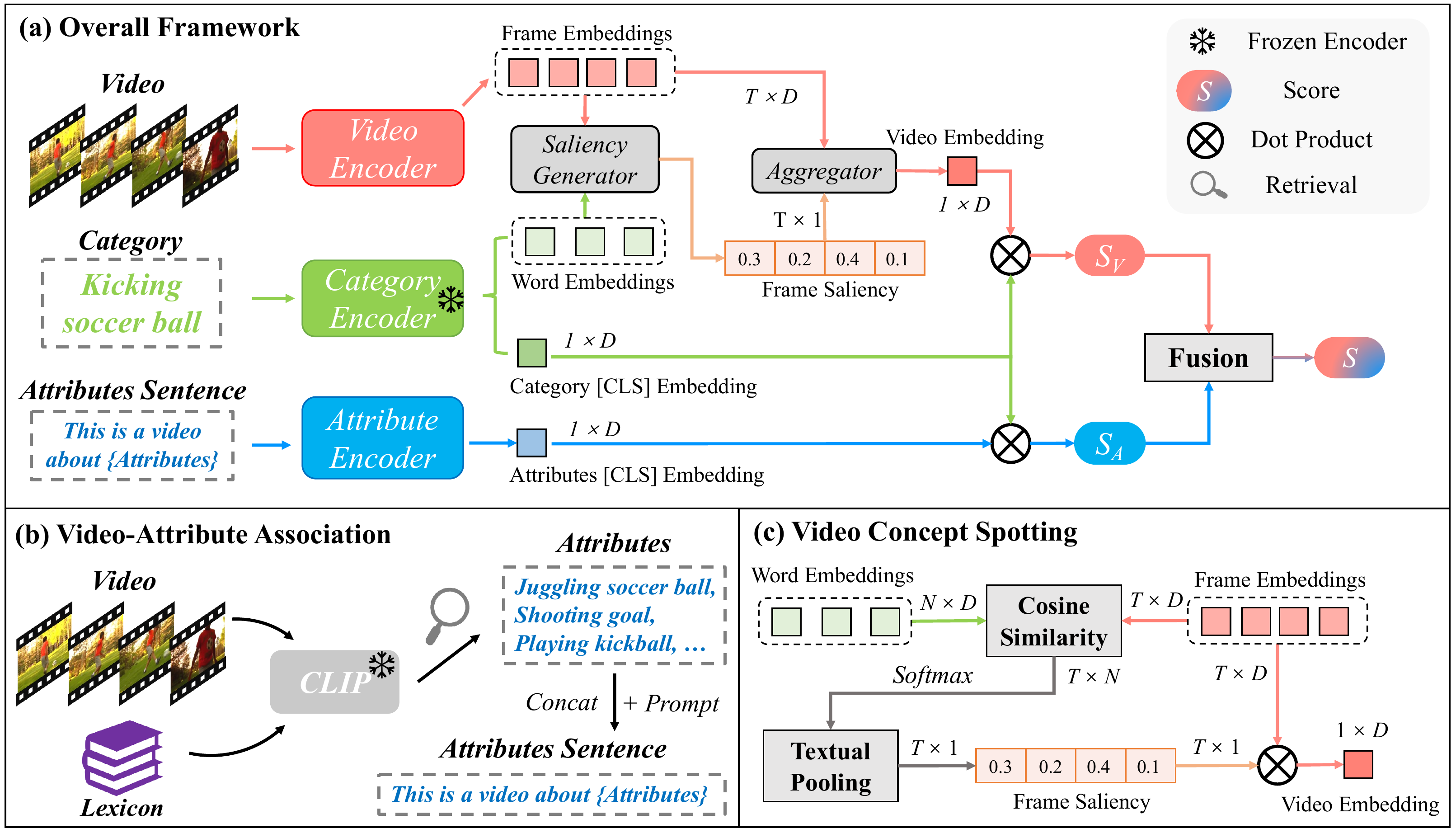}
\end{center}
\vspace{-1em}
\caption{An overview of our \textbf{BIKE} for video recognition. (a) \textbf{BIKE} explores bidirectional cross-modal knowledge from the pre-trained vision-language model (\eg, CLIP) to introduce auxiliary attributes and category-dependent temporal saliency for improved video recognition. \textbf{BIKE} comprises an auxiliary \emph{Attributes branch} and a main \emph{Video branch}.
(b) In the \texttt{Video-to-Text} direction, we present the \textbf{Video-Attribute Association} mechanism, which retrieves semantically relevant phrases from a pre-defined lexicon as video attributes for the input video. These attributes are concatenated and combined with a textual prefix to form an attribute sentence for text recognition.
(c) In the \texttt{Text-to-Video} direction, we present the \textbf{Video Concept Spotting} mechanism, which computes the similarity between video frames and a given category as a measure of temporal saliency to enhance video representation.
$D$ is the dimension of embedding, $T$ is the number of frames, and $N$ is the number of words in the category name.
}
\label{fig:approach}
\end{figure*}

An overview of our proposed \textbf{BIKE} is shown in Figure~\ref{fig:approach}. We next elaborate on each component in more detail.

\subsection{Preliminary: Video Recognition with VLM}
\label{sec:VL}
In this section, we describe the typical cross-modal video recognition pipeline~\cite{wang2021actionclip,ju2022prompting,x-clip,text4vis} based on the pre-trained vision-language model (VLM).
Given a video, we sample $T$ frames from the video as input $v$. 
We also have a collection of categories $C=\{c_1, c_2, \cdots, c_K\}$, where $K$ is the number of classes.
The goal of the video recognition task is to classify the video $v$ into a category $c \in C$.
Under the formulation of video recognition, the video $v$ is encoded with a vision encoder $f(\cdot | \theta_v)$ to obtain the video embedding $\mathbf{e_v}$, and the category $c$ is encoded with a text encoder $g(\cdot | \phi_c)$ to obtain the category embedding $\mathbf{e_c}$, where
\begin{equation}
    \mathbf{e_v} = f(v | \theta_v),  \mathbf{e_c} = g(c| \phi_c).
\end{equation}
Finally, we obtain the similarity score $\mathcal{S}_V$ as follows:
\begin{equation}
    \mathcal{S}_V = s(\mathbf{e_v},\mathbf{e_c}),
\end{equation}
where $s(\cdot,\cdot)$ is the cosine similarity function. The objective during training is to maximize $\mathcal{S}_V$ if $v$ and $c$ are matched, and minimize it in all other cases. During inference, we compute the score between the video embedding and each category embedding, and choose the category with the highest $\mathcal{S}_V$ as the top-1 prediction.
The parameter $\theta_v$ and $\phi_c$ of the video encoder and text encoder are initialized with weights from the pre-trained VLM (\eg, CLIP~\cite{CLIP}).
Throughout the rest of this work, we use the same notation.

\subsection{Video-to-Text: Video-Attributes Association }
\label{sec:VAL}
First we focus on exploring \texttt{Video-to-Text} auxiliary signals. We present an \emph{Attributes branch} as a complement to the regular \emph{Video branch} in \cref{sec:VL} for video recognition.

\noindent\textbf{Pre-generated Attributes.}
We begin by describing how to generate auxiliary attributes. As depicted in Figure~\ref{fig:approach}(b), we utilize the zero-shot capability of the VLM (\eg, CLIP~\cite{CLIP}) to identify the most relevant phases from a pre-defined lexicon as possible \emph{``Attributes"} of the video.
To achieve this, we first apply the CLIP's image encoder to the input video $V$ to extract frame-level features. These features are then combined using average pooling to yield a video embedding. Next, we feed each phase in the pre-defined lexicon into the CLIP's text encoder to produce a set of text embeddings.
We then calculate the similarity between this video embedding and each text embedding, sort the results, and select the top few phrases as the \emph{``Attributes"}. 
Once we have obtained the attributes, we employ a simple fusion method that concatenates them into a single attributes sentence $a$. We also add a manually-designed prompt as a prefix to the sentence, such as ``This is a video about \{\}''. 

\noindent\textbf{Attributes Recognition.} 
As shown in Figure~\ref{fig:approach}(a), the attributes sentence $a$ is encoded with a text encoder $g(\cdot | \phi_a)$ to produce the attribute embedding $\mathbf{e_a}$:
\begin{equation}
    \mathbf{e_a} = g(a| \phi_a).
\end{equation}
We use this attribute embedding to perform \emph{Attributes Recognition} by calculating the similarity $\mathcal{S}_A$ between the attribute embedding and category embeddings.
Note that both the attribute sentence and categories are encoded using the same text encoder from CLIP.
Interestingly, the Attributes branch can achieve a certain level of recognition performance (\eg, $ \sim $56\%) without any extra training, even though it's a lightweight text recognition pipeline.
During inference, we combine the well-trained Video branch with the plug-and-play Attributes branch using the following fusion equation:
\begin{equation}
    \mathcal{S} = \lambda \mathcal{S}_{V} + (1-\lambda) \mathcal{S}_{A},
    \label{eq:infer}
\end{equation}
where $\lambda$ is the fusion weight.
Without any additional training, the \emph{Attributes Recognition} surprisingly improve the video recognition performance, \eg, $78.8\%\xrightarrow{+1.2\%}80.0\%$ on the challenging Kinetics-400.
Naturally, the text encoder $g(\cdot | \phi_a)$ can be further trained in an end-to-end manner to improve the \emph{Attributes branch} and provide a stronger complementary capability, \eg, $78.8\%\xrightarrow{+2.6\%}81.4\%$.

\subsection{Text-to-Video: Video Concept Spotting}
\label{sec:sal}
In \cref{sec:VAL}, the \texttt{Video-to-Text} knowledge is employed to generate auxiliary attributes, thereby constructing a complementary \emph{Attributes branch}. 
Naturally, we also conduct an exploration to leverage the \texttt{Text-to-Video} knowledge to enhance the standard \emph{Video branch} for video recognition.
Specifically, we propose the use of category-dependent temporal saliency to guide the temporal aggregation process, resulting in a compact video representation that enhances video recognition.

\noindent\textbf{Background.}
To obtain a video representation based on a pre-trained image model, the typical pipeline involves two stages. First, we employ the image model to extract the spatial embedding of each frame. Next, the embeddings of these frames are temporally aggregated (\eg, mean pooling) to yield a video-level representation.

\begin{table*}[ht]
    \centering
    \scalebox{0.95}{
    \setlength{\tabcolsep}{2.0pt}
  \begin{tabular}{lcccccccc}
\toprule
  \textbf{Method} & \textbf{Venue} & \textbf{Input}   & \textbf{Pre-training} & \textbf{Top-1(\%)} & \textbf{Top-5(\%)} &  \textbf{Views} & \textbf{FLOPs} & \textbf{Param}  \\
  \midrule
  NL I3D-101~\cite{nonlocal} & CVPR'18 & 128\x224$^2$  & ImageNet-1K & 77.7 & 93.3 & 10\x3 & 359\x30 & 61.8\\
  MVFNet$_{En}$~\cite{wu2020MVFNet} & AAAI'21 & 24\x224$^2$  & ImageNet-1K & 79.1 & 93.8 & 10\x3 & 188\x30 & - \\
  
  TimeSformer-L~\cite{timesformer} & ICML'21 & 96\x224$^2$  & ImageNet-21K & 80.7 & 94.7 & 1\x3 & 2380\x3 & 121.4\\
  ViViT-L/16\x2~\cite{arnab2021vivit} & ICCV'21 & 32\x320$^2$  & ImageNet-21K & 81.3 & 94.7 & 4\x3 & 3992\x12 & 310.8\\  
  VideoSwin-L~\cite{videoswin} & CVPR'22 & 32\x384$^2$  & ImageNet-21K & 84.9 & 96.7 & 10\x5 & 2107\x50 & 200.0 \\  
  \hline
 \rowcolor{gray!20}
 \multicolumn{9}{l}{\emph{Methods with large-scale image pre-training}} \\  
  ViViT-L/16\x2~\cite{arnab2021vivit} & ICCV'21 & 32\x320$^2$ & JFT-300M & 83.5 & 95.5 &   4\x3 & 3992\x12 & 310.8\\
  ViViT-H/16\x2~\cite{arnab2021vivit} & ICCV'21 & 32\x224$^2$ & JFT-300M & 84.8 & 95.8 & 4\x3 & 8316\x12 & 647.5\\
  TokenLearner-L/10~\cite{ryoo2021tokenlearner} & NeurIPS'21 & 32\x224$^2$  & JFT-300M & 85.4 & 96.3 & 4\x3 & 4076\x12 & 450 \\
  MTV-H~\cite{MTV} & CVPR'22 & 32\x224$^2$   & JFT-300M & 85.8 & 96.6 & 4\x3 & 3706\x12 & - \\
  CoVeR~\cite{cover} & arXiv'21 & 16\x448$^2$  & JFT-300M & 86.3 & - & 1\x3 & - & - \\
  CoVeR~\cite{cover} & arXiv'21 & 16\x448$^2$  & JFT-3B & 87.2 & - & 1\x3 & - & - \\ 
  \hline
   \rowcolor{gray!20}
 \multicolumn{9}{l}{\emph{Methods with large-scale image-language pre-training}} \\
 CoCa ViT-giant~\cite{yu2022coca} & arXiv'22  & 6\x288$^2$ & JFT-3B+ALIGN-1.8B & 88.9 & - & - & - & 2100 \\ 
  VideoPrompt ViT-B/16~\cite{ju2022prompting} & ECCV'22 & 16\x224$^2$  & WIT-400M & 76.9 & 93.5 & - & - & - \\ 
  ActionCLIP ViT-B/16~\cite{wang2021actionclip} & arXiv'21  & 32\x224$^2$  & WIT-400M & 83.8 & 96.2 & 10\x3 & 563\x30 & 141.7 \\
  Florence~\cite{yuan2021florence} & arXiv'21  & 32\x384$^2$  & FLD-900M & 86.5 & 97.3 & 4\x3 & - & 647 \\   
  ST-Adapter ViT-L/14~\cite{st-adaptor} & NeurIPS'22 & 32\x224$^2$  & WIT-400M & 87.2 & 97.6 & 3\x1 & 8248 & - \\
   AIM ViT-L/14~\cite{yang2023aim} & ICLR'23 & 32\x224$^2$ & WIT-400M & 87.5 & 97.7 & 3\x1 & 11208 & 341 \\
  EVL ViT-L/14~\cite{evl} & ECCV'22 & 32\x224$^2$  & WIT-400M & 87.3 & - & 3\x1 & 8088 & - \\
  EVL ViT-L/14~\cite{evl} & ECCV'22 & 32\x336$^2$  & WIT-400M & 87.7 & - & 3\x1 & 18196 & - \\  
  X-CLIP ViT-L/14~\cite{x-clip} & ECCV'22 & 16\x336$^2$  & WIT-400M & 87.7 & 97.4 & 4\x3 & 3086\x12 & - \\
  Text4Vis ViT-L/14~\cite{text4vis} & AAAI'23 & 32\x336$^2$  & WIT-400M & 87.8 & 97.6 & 1\x3 & 3829\x3 & 230.7 \\
  \midrule
  \multirow{3}{*}{\textbf{BIKE ViT-L/14}} & \multirow{3}{*}{CVPR'23} & 16\x224$^2$  & \multirow{3}{*}{WIT-400M} & 88.1 & 97.9 & 4\x3 & 830\x12 & 230 \\ 
  & & 8\x336$^2$  &  & 88.3 & 98.1 & 4\x3 & 932\x12 & 230 \\
  &  & 32\x336$^2$  &  & \textbf{88.6} & \textbf{98.3} & 4\x3 & 3728\x12 & 230 \\ 
\bottomrule
  \end{tabular}   } 
 \caption{Comparisons with state-of-the-art methods on Kinetics-400. We report the FLOPs in inference phase.``Views'' indicates \# temporal clip $\times$ \# spatial crop. The magnitudes are Giga ($10^{9}$) and Mega ($10^{6}$) for FLOPs and Param. }
    \label{tab:k400_sota}
\end{table*}

\noindent\textbf{Parameter-Free Video Concept Spotting.} 
Mean pooling is a widely used technique to aggregate the frame embeddings and obtain the final video representation. 
Instead of treating each video frame equally as in mean pooling, we propose a parameter-free solution that utilizes the pre-aligned visual and textual semantics offered by the VLM (\eg, CLIP~\cite{CLIP}) to capture temporal saliency for video feature aggregation, as illustrated in Figure~\ref{fig:approach}(c). 
To estimate temporal saliency, we employ word embeddings as the query to obtain finer word-to-frame saliency.
Formally, the pre-trained VLM can encode each video or category name separately, and output two sets of embeddings: 
$\{\mathbf{v}_t \in \mathbb{R}^{d} | t=1,2,\cdots,T\}$ is a set of frame embeddings, where $T$ is the number of sampled frames, and $\{\mathbf{t}_n \in \mathbb{R}^{d}|n=1,2,\cdots,N\}$ is a set of word embeddings, where $N$ is the number of words in the class name.
We calculate the similarity between each word and each frame to measure the fine-grained relevancy. 
After that, we perform a softmax operation to normalize the similarities for each frame, and then aggregate the similarities between a certain frame and different words to obtain a frame-level saliency.

\begin{equation}
\mathcal{S}_{t}= \frac{1}{N} \sum_{n=1}^N  \dfrac{\exp({\mathbf{v}_t}^\mathsf{T}\mathbf{t}_n/\tau)}{\sum_{t=1}^T \exp({\mathbf{v}_t}^\mathsf{T} \mathbf{t}_n/\tau)}, t \in [1, T], n \in [1, N],
\label{eq:softmax}
\end{equation}
where $\tau$ is the temperature of this softmax function. 
\emph{See Figure~\ref{fig:vis} for the visualization of temporal saliency.}
Next, we utilize the temporal saliency to aggregate these frame embeddings as follows: 
\begin{equation}
    \mathbf{e_v} = \sum_{t=1}^T \mathbf{v}_t \mathcal{S}_{t},
\end{equation}
 $\mathbf{e_v} \in \mathbb{R}^{d}$ is the final enhanced video representation.

\subsection{Objectives of BIKE}
We present the \textbf{BIKE} learning framework for video recognition, as depicted in Figure~\ref{fig:approach}(a). 
Formally, our \textbf{BIKE} extracts feature representations $\mathbf{e_v}$, $\mathbf{e_a}$, and $\mathbf{e_c}$ for a given video $v$, pre-generated attributes $a$, and category $c$ with the corresponding encoders $f(\cdot | \theta_v)$, $g(\cdot | \phi_a)$, and $g(\cdot | \phi_c)$. 
Model parameters $\theta_v$, $\theta_a$, and $\theta_c$ are initialized with the weights from the pre-trained VLM (\eg, CLIP~\cite{CLIP}).
In this paper, we freeze the parameters of the pre-trained text encoder for $g(\cdot | \phi_c)$ and design extra manual prompts for the category $c$ and attributes sentence $a$. 

During the training phase, our objective is to ensure that the video representation $\mathbf{e_v}$ and the category representation $\mathbf{e_c}$ are similar when they are related and dissimilar when they are not, and the same applies to the attributes-category pairs. 
Given a batch of $B$ quadruples $\{\mathbf{e_v}_i, \mathbf{e_a}_i, \mathbf{e_c}_i \equiv C[y_i], y_i\}_{i=1}^{B}$, where $C$ is the collection of $K$ categories indexed by $y_i \in [0, K-1]$ and $y_i$ is a label indicating the index of the category in the dataset, and $\mathbf{e_v}_i$, $\mathbf{e_a}_i$, $\mathbf{e_c}_i$ denote the $i$-th video embedding, attributes embedding, and category embedding, respectively.
We follow the common practice \cite{wang2021actionclip,ju2022prompting} to consider the bidirectional learning objective and employ symmetric cross-entropy loss to maximize the similarity between matched \emph{Video-Category} pairs and minimize the similarity for other pairs: 
\begin{equation}
\begin{gathered}
\mathcal{L}_{V2C} =  - \frac{1}{B} \sum_{i}^{B} \frac{1}{ |\mathcal{K}(i)|  }  \sum_{ k \in \mathcal{K}(i) }
\log \frac{ \exp(s(\mathbf{e_c}_i,\mathbf{e_v}_k)/\tau)  }{\sum_{j}^{B}  \exp(s(\mathbf{e_c}_i,\mathbf{e_v}_j)/\tau) }, \\
\mathcal{L}_{C2V}	= - \frac{1}{B} \sum_{i}^{B}  \frac{1}{ |\mathcal{K}(i)|  }  \sum_{ k \in \mathcal{K}(i) }
\log \frac{ \exp(s(\mathbf{e_c}_k,\mathbf{e_v}_i)/\tau)  }{\sum_{j}^{B}  \exp(s(\mathbf{e_c}_j,\mathbf{e_v}_i)/\tau) }, \\
\mathcal{L}_{V} = \frac{1}{2} (\mathcal{L}_{V2C} + \mathcal{L}_{C2V}),
\end{gathered}    
\end{equation}
where $k \in \mathcal{K}(i) = \{ k | k \in [1, B], y_k = y_i\}$, $s(\cdot,\cdot)$ is the cosine similarity, and $\tau$ refers to the temperature hyper-parameter for scaling. 
Similarly, the loss for \emph{Attributes branch} is formulated as:
\begin{equation}
\begin{gathered}
\mathcal{L}_{A2C} =  - \frac{1}{B} \sum_{i}^{B} \frac{1}{ |\mathcal{K}(i)|  }  \sum_{ k \in \mathcal{K}(i) }
\log \frac{ \exp(s(\mathbf{e_c}_i,\mathbf{e_a}_k)/\tau)  }{\sum_{j}^{B}  \exp(s(\mathbf{e_c}_i,\mathbf{e_a}_j)/\tau) }, \\
\mathcal{L}_{C2A}	= - \frac{1}{B} \sum_{i}^{B}  \frac{1}{ |\mathcal{K}(i)|  }  \sum_{ k \in \mathcal{K}(i) }
\log \frac{ \exp(s(\mathbf{e_c}_k,\mathbf{e_a}_i)/\tau)  }{\sum_{j}^{B}  \exp(s(\mathbf{e_c}_j,\mathbf{e_a}_i)/\tau) }, \\
\mathcal{L}_{A} = \frac{1}{2} (\mathcal{L}_{A2C} + \mathcal{L}_{C2A}).
\end{gathered}    
\end{equation}
The total loss $\mathcal{L}$ is the sum of $\mathcal{L}_{V}$ and $\mathcal{L}_{A}$:
\begin{equation}
    \mathcal{L} = \mathcal{L}_{V} + \mathcal{L}_{A}.
\end{equation}
For inference, we simply combine the similarity score of the two branches as Equation \ref{eq:infer}.

\section{Experiments}
\subsection{Setups}
We conduct experiments on six widely used video benchmarks, \ie, Kinetics-400~\cite{kay2017kinetics} \& 600~\cite{k600}, ActivityNet~\cite{caba2015activitynet}, Charades~\cite{charades}, UCF-101~\cite{ucf101} and HMDB-51~\cite{hmdb}. 
\emph{See Supplementary for statistics of these datasets.}

\textbf{Training \& Inference.}\label{training}
In our experiments, we adopt the visual encoder of CLIP~\cite{CLIP} as the video encoder and use the textual encoder of CLIP for both the category and attributes encoders. To avoid conflict between the two branches, we first train the video encoder and then the attributes encoder.
To prepare the video input, we sparsely sample $T$ (\eg, 8, 16, 32) frames. We set the temperature hyperparameter $\tau$ to 0.01 for all training phases.
\emph{See Supplementary for detailed training hyperparameters.} 

To trade off accuracy and speed, we consider two evaluation protocols. 
(1) \emph{Single View}: We use only 1 clip per video and the center crop for efficient evaluation, as shown in Table~\ref{tab:ablations}.
(2) \emph{Multiple Views}: It is a common practice~\cite{slowfast,i3d,wu2020MVFNet} to sample multiple clips per video with several spatial crops to get higher accuracy. For comparison with SOTAs, we use four clips with three crops (``4\x3 Views'') in Table~\ref{tab:k400_sota}.

\begin{table*}[t]
    \begin{minipage}[t]{0.26\linewidth}
    \centering
    \scalebox{0.95}{
    \setlength{\tabcolsep}{2.0pt}
    \begin{tabular}{lcc} \toprule
      \textbf{Method}   & \textbf{Top-1} & \textbf{mAP} \\ \midrule
       ListenToLook~\cite{gao2020listentolook}  & - & 89.9 \\
       MARL~\cite{wu2019multi} & 85.7 & 90.1 \\ 
       DSANet~\cite{dsanet} & - & 90.5 \\ 
       TSQNet~\cite{tsqnet} & 88.7 & 93.7 \\ 
       NSNet~\cite{nsnet} & 90.2 & 94.3 \\        
       \midrule
       \textbf{BIKE ViT-L} & \textbf{94.7} & \textbf{96.1} \\ 
       \bottomrule
    \end{tabular}} 
     \caption{Comparisons with SOTAs on ActivityNet.}
    \label{tab:anet_sota}
    \end{minipage} 
    \hspace{2mm}
    \begin{minipage}[t]{0.3\linewidth}
    \centering
    \scalebox{0.95}{
    \setlength{\tabcolsep}{2.0pt}
    \begin{tabular}{lcc} \toprule
      \textbf{Method}   & \textbf{Frames} & \textbf{mAP} \\ \midrule
    MultiScale TRN \cite{trn} & - &  25.2 \\
    STM \cite{stm} & 16 &  35.3 \\
    SlowFast R101 \cite{slowfast} & 16+64 &  42.5 \\
    X3D-XL (312$\uparrow$) \cite{feichtenhofer2020x3d}  & 16 &  43.4 \\
    ActionCLIP \cite{wang2021actionclip} & 32 & 44.3 \\       
       \midrule
       \textbf{BIKE ViT-L} & \textbf{16} & \textbf{50.4} \\ 
       \bottomrule
    \end{tabular}} 
     \caption{Comparisons on Multi-label video dataset Charades.}
    \label{tab:charades}
    \end{minipage} 
    \hspace{2mm}
    \begin{minipage}[t]{0.41\linewidth}
    \centering
    \scalebox{0.95}{
    \setlength{\tabcolsep}{2.0pt}
    \begin{tabular}{lccccc} \toprule
      \textbf{Method}   & \textbf{Shot} & \textbf{HMDB} & \textbf{UCF} & \textbf{ANet} & \textbf{K400} \\ \midrule
       VideoSwin~\cite{videoswin} & 2 & 20.9  & 53.3  & - & - \\
       VideoPrompt~\cite{ju2022prompting}  & 5  & 56.6  & 79.5  & - & 58.5 \\
       X-Florence~\cite{x-clip}  & 2  & 51.6  & 84.0  & - & - \\ \midrule
       \multirow{3}{*}{\textbf{BIKE ViT-L}}
          & 1  & 72.3  & 95.2  & 86.6 & 73.5 \\
        & 2  & \textbf{73.5}  & \textbf{96.1}  &  \textbf{88.7} & \textbf{75.7} \\
        & 5  & \textbf{77.7}  & \textbf{96.5}  &  \textbf{90.9} & \textbf{78.2} \\
        \bottomrule
    \end{tabular} }   
    \caption{Comparisons on few-shot action recognition across four video datasets.}
    \label{table:video_few}
    \end{minipage}
\end{table*}

\begin{table*}[t]
    \centering
    \scalebox{0.95}{
    	\begin{tabular}{lllcc}
    	\toprule
    	\textbf{Method}  & \textbf{UCF$^*$ / UCF} & \textbf{HMDB$^*$ / HMDB} & \textbf{ActivityNet$^*$/ ActivityNet} & \textbf{Kinetics-600} \\ \midrule
    	GA~\cite{GA} & 17.3$\pm$1.1 / - & 19.3$\pm$2.1 / - & -  & -  \\
    	TS-GCN~\cite{TS-GCN} & 34.2$\pm$3.1 / - & 23.2$\pm$3.0 / - & - & - \\
        E2E~\cite{E2E} & 44.1 / 35.3 & 29.8 / 24.8 & 26.6 / 20.0 & - \\
        DASZL~\cite{DASZL} & 48.9$\pm$5.8 / - & - / - & - & -\\
        ER~\cite{ER} & 51.8$\pm$2.9 / - & 35.3$\pm$4.6 / - & - & 42.1$\pm$1.4 \\
        ResT~\cite{ResT} & 58.7$\pm$3.3 / 46.7 & 41.1$\pm$3.7 / 34.4 & 32.5 / 26.3 & - \\
        \midrule
        \textbf{BIKE ViT-L} & \textbf{86.6$\pm$3.4 / 80.8}  & \textbf{61.4$\pm$3.6 / 52.8}  & \textbf{86.2$\pm$1.0 / 80.0} & \textbf{68.5$\pm$1.2} \\
    	\bottomrule
    	\end{tabular} } 
 \caption{Comparisons on zero-shot video recognition. $^*$ denotes randomly selecting half of the test dataset's classes for evaluation, repeating the process ten times, and reporting the mean accuracy with standard deviation. For Kinetics-600, we adopt official code~\cite{ER} to select the 220 new categories outside of Kinetics-400 for evaluation.}
    \label{tab:sota_zero}        
\end{table*}

\subsection{Main Results}\label{sec:sota}
\noindent\textbf{Comparison with State-of-the-arts.}
We present our results on \textbf{Kinetics-400} in Table~\ref{tab:k400_sota} and compare our approach with SOTAs trained under various pre-training settings. 
Our approach outperforms regular video recognition methods while requiring significantly less computation, as shown in the upper table. 
We also demonstrate superiority over methods that use web-scale image pre-training, such as JFT-300M~\cite{JFT300M} and JFT-3B~\cite{JFT3B}. Our model performs better than all JFT-300M pre-trained methods, achieving a higher accuracy (+2.3\%) than CoVeR~\cite{cover}.
Surprisingly, our method even outperforms the JFT-3B pre-trained model (\textbf{88.6\%} \emph{v.s.} 87.2\%) despite the latter having almost 3 billion annotated images and a data scale 7.5\x larger than ours.
We further compare our method with others using web-scale image-language pre-training, such as CLIP~\cite{CLIP} and Florence~\cite{yuan2021florence}. Despite Florence having a larger dataset (2\x more data than the 400M image-text data used in CLIP), our approach still achieves a higher accuracy by 2.1\%.
Additionally, using only 8 frames and the same CLIP pre-training, our model performs on par with the best results of other methods, such as EVL~\cite{evl}, X-CLIP~\cite{x-clip}, and Text4Vis~\cite{text4vis}. 
When we use more frames as input, our method achieves a new state-of-the-art accuracy of 88.6\% under the CLIP pre-training setting.

We also evaluate our method on the untrimmed video dataset, \textbf{ActivityNet-v1.3}, to verify its generalizability. We fine-tune the Kinetics-400 pre-trained model with 16 frames, and report the top-1 accuracy and mean average precision (mAP) using the official evaluation metrics. Our approach significantly outperforms recent SOTAs, as shown in Table~\ref{tab:anet_sota}.
Furthermore, to demonstrate its effectiveness on smaller datasets, we also evaluate our method on \textbf{UCF-101} and \textbf{HMDB-51}, achieving top-1 accuracy of 98.8\% and 83.1\%, respectively. 
\emph{We include the results in the Supplementary due to space limitations.}

\begin{table*}
	\centering
		\begin{subtable}[th]{0.41\textwidth}
		\centering
		\scalebox{0.93}{
			\begin{tabular}{lcl}
			\toprule
			\textbf{Video branch} &  \textbf{$g(\cdot | \phi_c)$} & \textbf{Top-1(\%)} \\ \midrule
			Baseline: Mean Pool & \faUnlock  & 76.8 \\
			+ Video Concept Spotting  & \faUnlock  & 78.5 ({\color{teal}\textbf{+1.7}}) \\
			+ (Technique) Transf & \faUnlock  & 78.7 ({\color{teal}\textbf{+1.9}})  \\
			+ Frozen label encoder & \faLock & \textbf{78.9} ({\color{teal}\textbf{+2.1}}) \\
			\bottomrule
			\end{tabular}   }
			\caption{The effectiveness of temporal saliency. \faUnlock~means finetuning category encoder $g(\cdot | \phi_c)$. Transf is the temporal transformer.
			}
			\label{tab:ablation:video_sal}
		\end{subtable}	
	    \hspace{2mm}
		\begin{subtable}[th]{0.31\textwidth}
		\centering
		\scalebox{0.93}{
			\begin{tabular}{ccc}
			\toprule
		   \makecell[c]{\textbf{VCS} \\ \textbf{Source}} &	\makecell[c]{\textbf{Recognition} \\ \textbf{Source}}  & \textbf{Top-1}      \\ \midrule
		   Word Emb. & Word Emb.  & 78.1 \\
		   $[$CLS$]$ Emb. &  $[$CLS$]$ Emb. & 74.7 \\
		   Word Emb. &  $[$CLS$]$ Emb. & \textbf{78.5} \\
			\bottomrule
			\end{tabular}}
			\caption{Different category embeddings are used for Video Concept Spotting (VCS) and recognition.}
			\label{tab:ablation:query}
		\end{subtable}	
	    \hspace{2mm}
		\begin{subtable}[th]{0.23\textwidth}
		\centering
		\scalebox{0.93}{
		\setlength{\tabcolsep}{2.0pt}
			\begin{tabular}{ccc}
			\toprule
			\makecell[c]{\textbf{Attributes}}  & \makecell[c]{\textbf{Category}} & \textbf{Top-1}      \\ \midrule
		    \xmark & \xmark  & 46.2 \\
		    \cmark & \xmark  & 51.2  \\
			\cmark & \cmark  & \textbf{56.6}   \\
			\bottomrule
			\end{tabular}}
			\caption{The effects of the textual prompt in \emph{Attributes recognition branch}  (w/o training).}
			\label{tab:ablation:prompt}
		\end{subtable}		    
	    \\[7pt]
		\begin{subtable}[th]{0.2\textwidth}
		\centering
		\scalebox{0.93}{
		\setlength{\tabcolsep}{2.0pt}
			\begin{tabular}{ccc}
			\toprule
			\textbf{\#Attributes}  & \textbf{A} & \textbf{V+A}     \\ \midrule
		    3 & 53.4  & 79.9  \\
			5 & 56.6 & \textbf{80.0}    \\
			7 & 57.1 & 79.7    \\
			\bottomrule
			\end{tabular}}
			\caption{Study on different number of attributes (w/o training).}
			\label{tab:ablation:n_attr}
		\end{subtable}	
		\hspace{2mm}
		\begin{subtable}[th]{0.28\textwidth}
		\centering
 		\scalebox{0.93}{
 		\setlength{\tabcolsep}{2.0pt}
			\begin{tabular}{ccc}
			\toprule
		  \textbf{Training} & \textbf{A} & \textbf{V $\xrightarrow{{\color{teal}\textbf{+$\Delta$\%}}}$ V+A}     \\ \midrule
		  \xmark  & 56.6 & 78.9 $\xrightarrow{{\color{teal}\textbf{+1.1\%}}}$ 80.0 \\ 
		  \cmark  & \textbf{69.6} & \textbf{78.9} $\xrightarrow{{\color{teal}\textbf{+2.5\%}}}$ \textbf{81.4} \\
			\bottomrule
			\end{tabular}}
			\caption{The impact of \emph{Attributes branch}. \cmark~means fine-tuning the attributes encoder.}
			\label{tab:ablation:train_attr}
		\end{subtable}	
		\hspace{2mm}
		\begin{subtable}[th]{0.23\textwidth}
		\centering
		\scalebox{0.93}{
		\setlength{\tabcolsep}{2.0pt}
			\begin{tabular}{lc}
			\toprule
		    	& \textbf{V $\xrightarrow{{\color{teal}\textbf{+$\Delta$\%}}}$ V+A} \\ \midrule
			Baseline &  76.8 $\xrightarrow{{\color{teal}\textbf{+2.4\%}}}$ 79.2 \\
			Ours  & 78.9 $\xrightarrow{{\color{teal}\textbf{+2.5\%}}}$ \textbf{81.4} \\
			\bottomrule
			\end{tabular}}
			\caption{The effects of \emph{Attributes branch} to complement \emph{Video branch}.}
			\label{tab:ablation:val}
		\end{subtable}	 		
		\hspace{2mm}
		\begin{subtable}[th]{0.21\textwidth}
		\centering
        \scalebox{0.93}{
		\setlength{\tabcolsep}{2.0pt}
			\begin{tabular}{lc}
			\toprule
			\textbf{Lexicon} & \textbf{V $\xrightarrow{{\color{teal}\textbf{+$\Delta$\%}}}$ V+A} \\ \midrule
			IN-1K &  78.9 $\xrightarrow{{\color{teal}\textbf{+1.4\%}}}$ 80.3 \\
			K400  & 78.9 $\xrightarrow{{\color{teal}\textbf{+2.5\%}}}$ \textbf{81.4} \\	
			\bottomrule
			\end{tabular}}
			\caption{Study on the impact of different lexicon.}
			\label{tab:ablation:lexicon}
		\end{subtable}			
		\\[7pt]
		\begin{subtable}[th]{0.44\textwidth}
		\centering
        \scalebox{0.93}{
			\begin{tabular}{lccl}
			\toprule
			\textbf{Method}    & \textbf{$T$} &  \textbf{Backbone} & \textbf{Top-1(\%)}    \\ \midrule
			VideoPrompt~\cite{ju2022prompting} & 16 & ViT-B/32 & 76.9 \\
			ActionCLIP~\cite{wang2021actionclip} & 8 & ViT-B/32 & 78.4 \\
            BIKE (Ours)  & 8 & ViT-B/32 & \textbf{81.4} ({\color{teal}\textbf{+3.0}})  \\
            \bottomrule
			\end{tabular}}
			\caption{Comparison with CLIP-based methods using single-view inference. $T$ is the number of frames.}
			\label{tab:ablation:clip-based}
		\end{subtable}			
		\hspace{2mm}
		\begin{subtable}[th]{0.53\textwidth}
		\centering
		\scalebox{0.93}{
			\begin{tabular}{lcc}
			\toprule
			\textbf{Backbone}  & \textbf{Baseline $\rightarrow$ V $\rightarrow$ V+A} & \textbf{V$^*$ $\rightarrow$ V$^*$+A} \\ \midrule
			ViT-B/32  & 76.8 $\xrightarrow{{\color{teal}\textbf{+2.1\%}}}$ 78.9 $\xrightarrow{{\color{teal}\textbf{+2.5\%}}}$ 81.4 &  80.2 $\xrightarrow{{\color{teal}\textbf{+1.7\%}}}$ 81.9 \\
            ViT-B/16  & 79.9 $\xrightarrow{{\color{teal}\textbf{+2.2\%}}}$ 82.1 $\xrightarrow{{\color{teal}\textbf{+1.1\%}}}$ 83.2 & 83.2 $\xrightarrow{{\color{teal}\textbf{+0.7\%}}}$ 83.9 \\
            ViT-L/14  & 85.2 $\xrightarrow{{\color{teal}\textbf{+0.8\%}}}$ 86.4 $\xrightarrow{{\color{teal}\textbf{+0.1\%}}}$ 86.5 & 87.4 $\xrightarrow{\textbf{+0\%}}$ 87.4  \\
            \bottomrule
			\end{tabular}}
			\caption{Component-by-component evaluation of our approach using various backbones. Models are fed 8 frames, where * stands for multiple view inference.}
			\label{tab:ablation:backbone}
		\end{subtable}	
		\caption{Ablation studies on Kinetics-400. Models use ViT-B/32 as the backbone, and 8 frames as input, unless otherwise specified. We report top-1 accuracy (\%) for a single clip input with 224$\times$224 spatial size. The \textbf{V} and \textbf{A} abbreviations are used for the \emph{Video recognition branch} and \emph{Attributes recognition branch}, respectively. We refer to ImageNet-1K and Kinetics-400 as IN-1K and K400, respectively.
		}
		\label{tab:ablations}
	\end{table*}

\noindent\textbf{Multi-Label Video Recognition.}\label{exp:ml_video}
In addition to the single-label video recognition, we also evaluate our method on multi-label video recognition. We use the \textbf{Charades} dataset, which contains long-term activities with multiple actions, and utilize the Kinetics-400 pre-trained ViT-L backbone for training. The results are evaluated using the mAP metric. As shown in Table~\ref{tab:charades}, our BIKE achieves the performance of 50.4\% mAP, demonstrating its effectiveness in multi-label video classification.

\noindent\textbf{Few-Shot Video Recognition.}\label{exp:video_fewshot}
We demonstrate the few-shot recognition capability of our method, which refers to video recognition using only a few training samples. In this experiment, we scaled up the task to categorize all categories in the dataset with only a few samples per category for training. We used a CLIP pre-trained ViT-L/14 with 8 frames for few-shot video recognition, without further Kinetics-400 pre-training. The top-1 accuracy on four datasets is reported in Table~\ref{table:video_few}. Our method shows remarkable transferability to diverse domain data in a data-poor situation.
On UCF-101 and HMDB-51, our method outperforms VideoSwin~\cite{videoswin} by 42.8\% and 52.6\%, respectively. In comparison with image-language pre-trained methods, our method outperforms VideoPrompt~\cite{ju2022prompting} and X-Florence~\cite{x-clip} by 21.1\% and 21.9\% on HMDB-51, respectively. 
\emph{See Supplementary for training details.}

\noindent\textbf{Zero-shot Video Recognition.}\label{exp:zero_video}
We further evaluate our method in an open-set setting. Table~\ref{tab:sota_zero} presents the results of zero-shot evaluation on four video datasets using our Kinetics-400 pre-trained model (\ie, ViT-L/14 with 8 frames).
There are two major evaluation methods on UCF-101, HMDB-51, and ActivityNet: half-classes evaluation (marked as $^*$) and full-classes evaluation. For fair comparison, we present the results under the half-classes evaluation protocol, which has been widely used in previous works~\cite{GA,E2E,ER,ResT}. Additionally, we provide results on the entire dataset for more challenging and realistic accuracy evaluation.
\emph{See Supplementary for further details on evaluation protocols.}
Our method exhibits strong cross-dataset generalization ability and outperforms classic zero-shot video recognition methods. 



\subsection{Ablation Studies}
\label{ab:k400}
In this section, we provide extensive ablations to demonstrate our method with the instantiation in Table~\ref{tab:ablations}.

\noindent\textbf{The Effect of Temporal Saliency.}
We investigate the impact of our proposed \emph{Video Concept Spotting} (VCS) mechanism on the performance of the \emph{Video branch}, as shown in Table~\ref{tab:ablation:video_sal}. We start with a baseline that uses mean pooling to aggregate the features of all frames, without considering temporal saliency. We observe that equipping the baseline with VCS can improve the accuracy by +1.7\%.
We then introduce a multi-layer (\eg, 6-layers) Transformer encoder with position embedding for sequence features, commonly used in previous methods, and find that it provides an additional 0.2\% performance boost. Moreover, freezing the category encoder not only reduces training parameters but also slightly improves performance (+0.2\%).

\noindent\textbf{Exploration of Category Embedding for Temporal Saliency and Classification.} 
As mentioned in Section \ref{sec:sal}, CLIP's textual encoder can generate two types of embeddings: the [CLS] embedding for the entire sentence and the word embedding for each word.
Therefore, we can encode the category into these two types of embeddings.
The category embedding has two roles in our method: 1) it serves as a query to determine the temporal saliency, and 2) it calculates similarity with video representation to produce recognition results.
We demonstrate the results for three different combinations in Table~\ref{tab:ablation:query}.
We find that the global [CLS] embedding performs better than the word-level embedding for final recognition, but the word-level embedding is necessary for temporal saliency.

\noindent\textbf{Prompt Engineering and Number of Attributes.}
For both attributes and categories in the \emph{Attributes recognition} branch, we manually define a prompt, \ie, ``This is a video about \{\}''. The results in Table~\ref{tab:ablation:prompt} show that the prompt significantly improves accuracy, even without training the attributes encoder. Furthermore, in Table~\ref{tab:ablation:n_attr}, we observe that the number of attributes has little effect on the performance of the \emph{Attributes recognition} and two-branch recognition.

\noindent\textbf{The Impact of \emph{Attributes branch}.} 
Table~\ref{tab:ablation:train_attr} shows that without any training, the \emph{Attributes branch} can be plug-and-played on the \emph{Video branch} to improve the recognition performance.
After training the attributes encoder, the \emph{Attributes branch} further boosts performance by an impressive 2.5\% on the fusion result. Additionally, we find that the \emph{Attributes branch} can also improve the baseline when fused with it, as shown in Table~\ref{tab:ablation:val}. By combining the \emph{VCS} and the \emph{Attributes branch}, we can achieve a remarkable improvement of 4.6\% on the baseline.

\noindent\textbf{Attributes Generation with Different Lexicons.}
In \cref{sec:VAL}, we use a pre-defined lexicon to obtain attributes. In Table~\ref{tab:ablation:lexicon}, we explore the impact of different lexicons. We used ImageNet-1K, an image dataset that covers 1000 object categories, as our lexicon to search for potential object attributes. According to the results, this can increase the performance of the \emph{Attributes branch} by 1.4\%. We found that using the 400 categories of Kinetics-400 as the lexicon can further improve the results.

\noindent\textbf{Comparison with CLIP-Based Methods.}
Table~\ref{tab:ablation:clip-based} presents a comparison between our method and two CLIP-based approaches, VideoPrompt~\cite{ju2022prompting} and ActionCLIP~\cite{wang2021actionclip}, both trained with contrastive loss.
Despite using fewer frames, our method achieves higher Top-1 accuracy than VideoPrompt. Moreover, using the same ViT-B/32 backbone, our approach outperforms ActionCLIP by \textbf{3.0\%}.

\noindent\textbf{More Evaluation with Different Backbones.} 
Table~\ref{tab:ablation:backbone} presents a comprehensive evaluation of the applicability of our method using larger backbones. Our observations are as follows:
1) Despite the greatly improved performance of the baseline with larger backbones, our \emph{VCS} mechanism still provides consistent, additional gains. This demonstrates the continued necessity of \texttt{Text-to-Video} saliency knowledge for large models.
2) As the absolute accuracy of the \emph{Video branch} increases, the complementing effect of the \emph{Attributes branch} gradually weakens. We conjecture that with larger models, richer representations are learned, leading to reduced bias in learned representations and an increased correlation with the \emph{Attributes branch}, resulting in a reduction in complementary information.
3) Multiple-view evaluation involving more video clips leads to increased performance, reducing the bias of the model itself. For models with a top-1 accuracy of 87.4\%, the \emph{Attributes branch} is unable to provide supplementary knowledge. Therefore, the \emph{Attributes branch} is not utilized in our ViT-L/14 models presented in \cref{sec:sota}.



\subsection{Visualization}
\begin{figure}[ht]
\begin{center}
\includegraphics[width=1\linewidth]{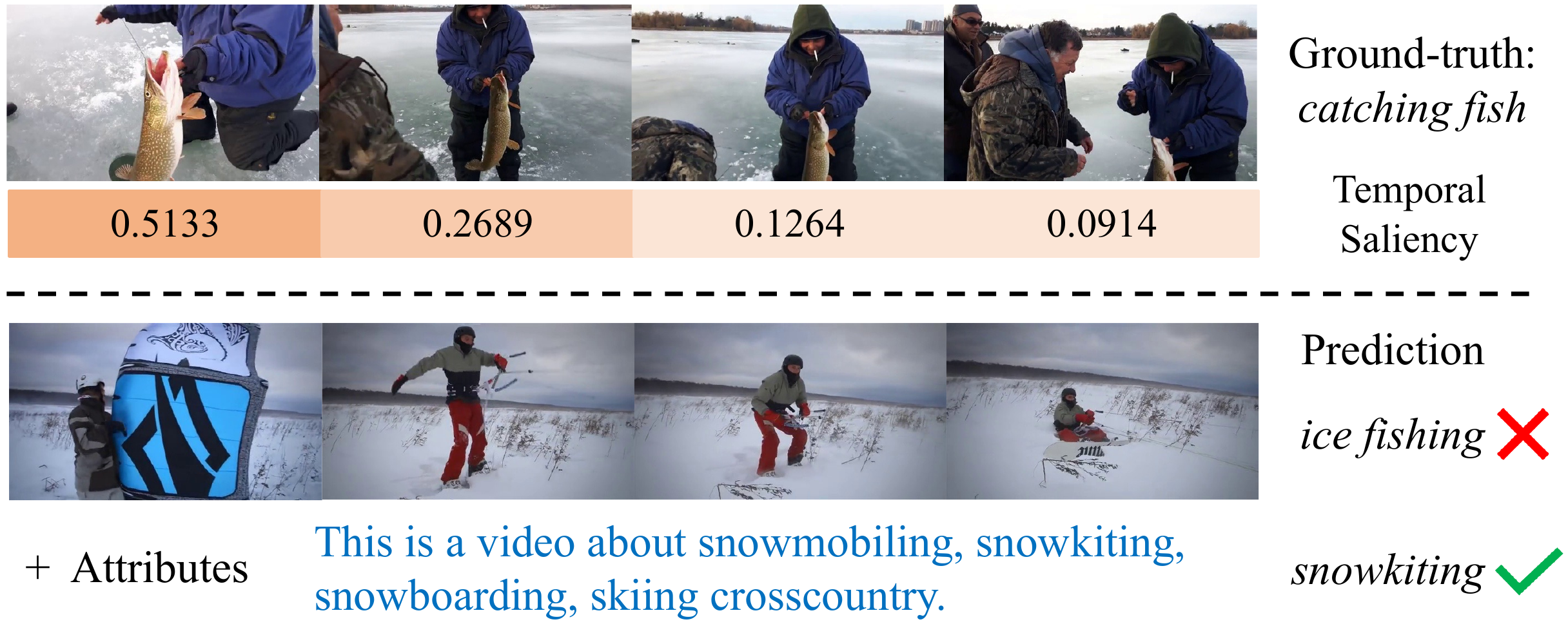}
\end{center}
\vspace{-1mm}
\caption{Visualization of (\textbf{Top}) temporal saliency and (\textbf{Bottom}) attributes. Please zoom in for the best view.
}
\label{fig:vis}
\end{figure}

Figure~\ref{fig:vis} illustrates the temporal saliency generated by Video Concept Spotting mechanism, highlighting the frame that is most relevant to the category. We also demonstrate the complementarity of the auxiliary attributes generated by our Video-Attribute Association mechanism with the video branch.
\emph{See more qualitative results in Supplementary.}

\section{Related Works}
\paragraph{Video Recognition.} 
Convolutional networks have been the standard backbone architecture in video recognition for a long time. Early works focused on jointly learning spatial and temporal context through parallel branches~\cite{two-stream,two-stream-stresnet,tsn,slowfast,wang2020symbiotic,wang2021interactive}. Later works developed plug-and-play temporal modules~\cite{wu2020MVFNet,dsanet,teinet,li2020tea,liu2021tam,p3d,s3d,r2+1d,tdn} for 2D CNN backbones to improve temporal modeling. Some works also designed dynamic inference mechanisms~\cite{nsnet,tsqnet,wu2020dynamic,wu2019multi,adafocus,wang2023differentiable} for efficient video recognition. Recently, Vision Transformers~\cite{ViT,DeiT,liu2021swin} has emerged as a new trend in image recognition backbones. Transformers have also been adopted for video recognition, such as TimeSFormer~\cite{timesformer}, ViViT~\cite{arnab2021vivit}, VideoSwin~\cite{videoswin}, and MViT~\cite{mvit}.

\noindent\textbf{Transferring CLIP Models for Video Recognition.}
CLIP~\cite{CLIP} provides good practice in learning the coordinated vision-language models using large-scale image and text pairs. The pre-trained model can learn powerful visual representations aligned with rich linguistic semantics. 
Initially, some works~\cite{luo2022clip4clip,zhao2022centerclip,wu2022cap4video,fang2023uatvr} propose to directly use CLIP for video-text retrieval. 
Later, a few works also explore the use of CLIP models for video recognition~\cite{wang2021actionclip,ju2022prompting,evl,x-clip,st-adaptor,text4vis,yang2023aim}, they can be broadly categorized into two lines. The first line~\cite{st-adaptor,evl,yang2023aim} follows the unimodal transferring paradigm, where the image encoder of CLIP is used as a strong initialization for the video encoder. The second line~\cite{wang2021actionclip,ju2022prompting,text4vis,x-clip} provides cross-model learning baselines that directly extend CLIP to video-label matching for video recognition.
However, these studies only briefly tap into the knowledge from CLIP.
In contrast, our work aims to further explore the bidirectional cross-modal knowledge from CLIP to enhance the cross-model baseline. Our approach introduces auxiliary attributes in the \texttt{Video-to-Text} direction and category-dependent temporal saliency in the \texttt{Text-to-Video} direction, resulting in a more effective and interpretable video recognition.

\section{Conclusion}


In this work, we introduce a novel two-stream framework called \textbf{BIKE} that leverages bidirectional cross-modal knowledge from CLIP models to enhance video recognition. Our approach involves the \emph{Attributes branch}, which utilizes the \textbf{Attributes-Category Association} mechanism to generate attributes for auxiliary recognition, and the \emph{Video branch}, which uses the \textbf{Video Concept Spotting} mechanism to generate temporal saliency and produce a more compact video representation. Our approach is evaluated on six video datasets, and the experimental results demonstrate its effectiveness.

{\small
\bibliographystyle{ieee_fullname}
\bibliography{egbib}
}

\clearpage
\twocolumn[\begin{center}\textbf{\Large Bidirectional Cross-Modal Knowledge Exploration for Video Recognition \\ with Pre-trained Vision-Language Models \vspace{2mm}
\\ \textnormal{\emph{Supplementary Material}}}\end{center}
\vspace{1em}
\begin{center}
\large Wenhao Wu$^{1,2}$\quad
Xiaohan Wang$^{3}$\quad
Haipeng Luo$^{4}$\quad
Jingdong Wang$^{2}$\quad
Yi Yang$^{3}$\quad
Wanli Ouyang$^{5,1}$\\
$^1$The University of Sydney \qquad $^2$Baidu Inc. \qquad $^3$Zhejiang University \\ 
$^4$University of Chinese Academy of Sciences  \qquad $^5$Shanghai AI Laboratory\\
{\tt\small whwu.ucas@gmail.com}
\vspace{2.5em}
\end{center}]

\appendix
\setcounter{table}{0}
\setcounter{figure}{0}
\renewcommand{\thetable}{A.\arabic{table}}
\renewcommand{\thefigure}{A.\arabic{figure}}

In this appendix, we provide additional details and results for our approach.
Specifically, \S\ref{sec:imp_detail} contains further \textit{details} on the training process (\S\ref{sec:train_video}), attributes branch (\S\ref{sec:attr}), zero-shot evaluation (\S\ref{sec:zero}), statistics of video datasets (\S\ref{sec:video_datasets}), visual encoder architectures (\S\ref{sec:vis_encoder}), and Distributed InfoNCE (\S\ref{sec:batch_gather}). 
In \S\ref{supp:video}, we present additional \textit{results}, including comparisons on UCF-101 and HMDB-51 (\S\ref{sec:small_sota}) and more visualizations (\S\ref{sec:vis}).

\section{Implementation Details}\label{sec:imp_detail}
\subsection{Training details}\label{sec:train_video}
\noindent\textbf{Regular Video Recognition}. 
We present our approach for regular video recognition in Table~\ref{tb:imp}, sharing the same training recipe for all video datasets, including Kinetics-400, ActivityNet, Charades, HMDB-51, and UCF-101.

\noindent\textbf{Few-shot Video Recognition}.
We repeat the samples to maintain the same number of iterations as the regular counterpart. For instance, if the model is trained on Kinetics-400 with around 900 iterations per epoch for the general setting, we repeat the sample to maintain the same number of iterations for few-shot settings. We train few-shot models for 2 epochs on Kinetics-400 and 10 epochs on other video datasets, \ie, ActivityNet, HMDB-51, and UCF-101, while keeping other settings the same as in Table~\ref{tb:imp}.

\noindent\textbf{Zero-shot Video Recognition}. 
We use the Kinetics-400 pre-trained models to perform cross-dataset recognition without additional training on other datasets such as ActivityNet, HMDB-51, UCF-101, and Kinetics-600.

\subsection{Attributes Branch}\label{sec:attr}
To improve the quality of auxiliary attributes, we pre-generate them using CLIP ViT-L/14 with 8 frames. We employ the text encoder architecture of CLIP ViT-B/32 as our attribute encoder.
To integrate the \emph{Attributes branch} with the \emph{Video branch}, we set $\lambda$ to 0.6 for the \emph{Video branch} with ViT-B and $\lambda$ to 0.8 for the \emph{Video branch} with ViT-L.

\begin{table}[h!]
\centering
\scalebox{0.95}{
\begin{tabular}{l|c} \toprule
    Setting  & Value \\ \midrule
    \multicolumn{2}{l}{\cellcolor{mygray}\emph{Training Hyperparameter}} \\
    Batch size & 256 \\
    Vocabulary size & 49408 \\
    Training epochs & 30 (ViT-B), 20 (ViT-L) \\
    Optimizer & AdamW \\
    Learning rate (Base) & 5e-5, cosine \\
    Learning rate (CLIP layers) & 5e-6, cosine  \\
    Weight decay & 0.2 \\
    Linear warm-up epochs & 5 \\
    Adam $\beta_1$,$\beta_2$ & 0.9, 0.999 \\ \midrule
    \multicolumn{2}{l}{\cellcolor{mygray}\emph{Augmentation}} \\
    Resize & RandomSizedCrop \\
    Crop size & 224 (Default) \\
    Random Flip & 0.5 \\
    Random Gray scale & 0.2 \\
    \bottomrule
\end{tabular}}
\caption{Default training recipe for video recognition.}
\label{tb:imp}
\end{table}

\begin{table*}[h]
\centering
\begin{tabular}{l|cccccccc} \toprule
           & Embedding & Input      & \multicolumn{3}{c}{Vision Transformer} & \multicolumn{3}{c}{Text Transformer} \\
    Model  & dimension & resolution & layers & width & heads  & layers & width & heads \\ \midrule
    ViT-B/32 & 512 & 224 & 12 & 768 & 12 & 12 & 512 & 8 \\
    ViT-B/16 & 512 & 224 & 12 & 768 & 12 & 12 & 512 & 8 \\
    ViT-L/14 & 768 & 224 & 24 & 1024 & 16 & 12 & 768 & 12 \\
    ViT-L/14-336px  & 768 & 336 & 24 & 1024 & 16 & 12 & 768 & 12 \\
    \bottomrule
\end{tabular}
\caption{CLIP-ViT hyperparameters}
\label{vit}
\end{table*}

\subsection{Evaluation Protocols for Zero-shot Recognition}\label{sec:zero}
We employ our Kinetics-400 pre-trained models to evaluate on other datasets. For UCF-101, HMDB-51, and ActivityNet, we adopt two major evaluation protocols as described in \cite{E2E}:
\begin{enumerate}
\item \textbf{Half-Classes Evaluation:} To ensure comparability with previous works, we randomly select half of the test dataset's classes - 50 for UCF, 25 for HMDB, and 100 for ActivityNet - and evaluate on the selected subset. We repeat this process ten times and average the results for each test dataset. We refer to this setting as UCF$^*$, HMDB$^*$ and ActivityNet$^*$.
\item \textbf{Full-Classes Evaluation:} This evaluation setting involves directly evaluating on the full dataset to return more realistic accuracy scores.
\end{enumerate}

For Kinetics-600, we follow \cite{ER} to choose the 220 new categories outside of Kinetics-400 in Kinetics-600 for evaluation. We use the three splits provided by \cite{ER} and sample 160 categories for evaluation from the 220 categories in Kinetics-600 for each split. We report the mean accuracy of the three splits as the final accuracy.

\subsection{Statistics of Video Datasets}\label{sec:video_datasets}
We describe the video datasets used in our experiments:

\noindent\textbf{Kinetics-400} is a large-scale video dataset that includes 240,000 training videos and 20,000 validation videos across 400 different human action categories. Each video in the dataset is a 10-second clip of an action moment, annotated from raw YouTube videos.

\noindent\textbf{Kinetics-600} is an extension of Kinetics-400, consisting of approximately 480,000 videos from 600 action categories. The videos are divided into 390,000 for training, 30,000 for validation, and 60,000 for testing. In this paper, we use its test set for zero-shot evaluation.

\noindent\textbf{UCF-101} is an action recognition dataset that contains 13,320 videos from 101 realistic action categories, collected from YouTube.

\noindent\textbf{HMDB-51} is a collection of realistic videos from various sources, including movies and web videos. The dataset comprises 7,000 video clips from 51 action categories.

\noindent\textbf{ActivityNet-v1.3} is a large-scale untrimmed video benchmark that contains 19,994 untrimmed videos of 5 to 10 minutes from 200 activity categories.

\noindent\textbf{Charades} is a video dataset designed for action recognition and localization tasks. It contains over 10,000 short video clips of people performing daily activities, and consists of 157 action categories.

\subsection{Encoder Architectures}\label{sec:vis_encoder}
In this paper, we provide the complete architecture details of the visual encoder and textual encoders. The CLIP-ViT architectures are shown in Table~\ref{vit}.

\begin{algorithm*}[h]
	\caption{Numpy-like Pseudocode of Distributed InfoNCE for our \emph{Video branch}}
	\label{batch_gather}
	\begin{lstlisting}[language=python]
    # category_encoder: encoder network for category input
    # video_encoder: encoder network for video input
    # V: minibatch of video inputs
    # T: minibatch of category inputs
    # N: the local batch size of each GPU, e.g.,16
    # M: the number of GPUs, e.g.,8
    # N * M: the global batch size for multi-gpu training, e.g.,128
    
    # extract feature representations of each modality
    local_vision_features = video_encoder(V) # shape: [N, embed_dim]
    local_text_features = category_encoder(T) # shape: [N, embed_dim]

    # normalization
    local_vision_features = l2_normalize(local_vision_features, axis=1)
    local_text_features = l2_normalize(local_text_features, axis=1)
    
    # batch_gather is a function gathering and concatenating the tensors across GPUs. 
    all_vision_features = batch_gather(local_vision_features) # shape: [N * M, embed_dim]
    all_text_features = batch_gather(local_text_features) # shape: [N * M, embed_dim]
    
    # scaled pairwise cosine similarities
    # shape = [N, N * M]
    logits_per_vision = logit_scale * local_vision_features @ all_text_features.t()  
    # shape = [N, N * M]
    logits_per_text = logit_scale * local_text_features @ all_vision_features.t() 
    
    # The logits are then used as inputs for N*M-way (e.g., 128-way) classification, 
    # resulting in a loss value corresponding to N inputs in each GPU. 
    # Then Distributed Data Parallel mechanism takes care of averaging these across GPUs, 
    # which becomes equivalent to calculating the loss over NMxNM (e.g.,128x128) similarities.
	\end{lstlisting}
\end{algorithm*}

\subsection{Distributed InfoNCE}\label{sec:batch_gather}
\label{sm:batch_gather}
Instead of Data-Parallel Training (DP), which is single-process, multi-thread, and only works on a single machine, Distributed Data-Parallel Training (DDP) is a widely adopted single-program multiple-data training paradigm for single- and multi-machine training.
Due to GIL contention across threads, per-iteration replicated model, and additional overhead introduced by scattering inputs and gathering outputs, DP is usually slower than DDP even on a single machine.
Hence, we develop the Distributed InfoNCE based on DDP for large batch size and fast training.

The core of the Distributed InfoNCE implementation is batch gathering, which enables us to calculate the NM\x NM similarity matrix across M GPUs for InfoNCE loss. Without batch gathering, each GPU only computes a local N\x N matrix where N$\ll$NM. This means that the cosine similarity and the InfoNCE loss would only be calculated for the pairs within a single GPU, and their gradients would be later averaged and synced. That's obviously not what we want.

The batch gathering technique allows each GPU to hold N vision features and perform a matrix product with NM text features, resulting in an N\x NM matrix. This computation is distributed (\ie, sharded) across M GPUs, and we have calculated NM\x NM similarities across the GPUs in total. The loss we employ is symmetric, and the same process is applied \emph{w.r.t.} text inputs.
\Cref{batch_gather} provides an example pseudocode to help understand the process.

\begin{figure*}
\begin{center}
\includegraphics[width=0.8\textwidth]{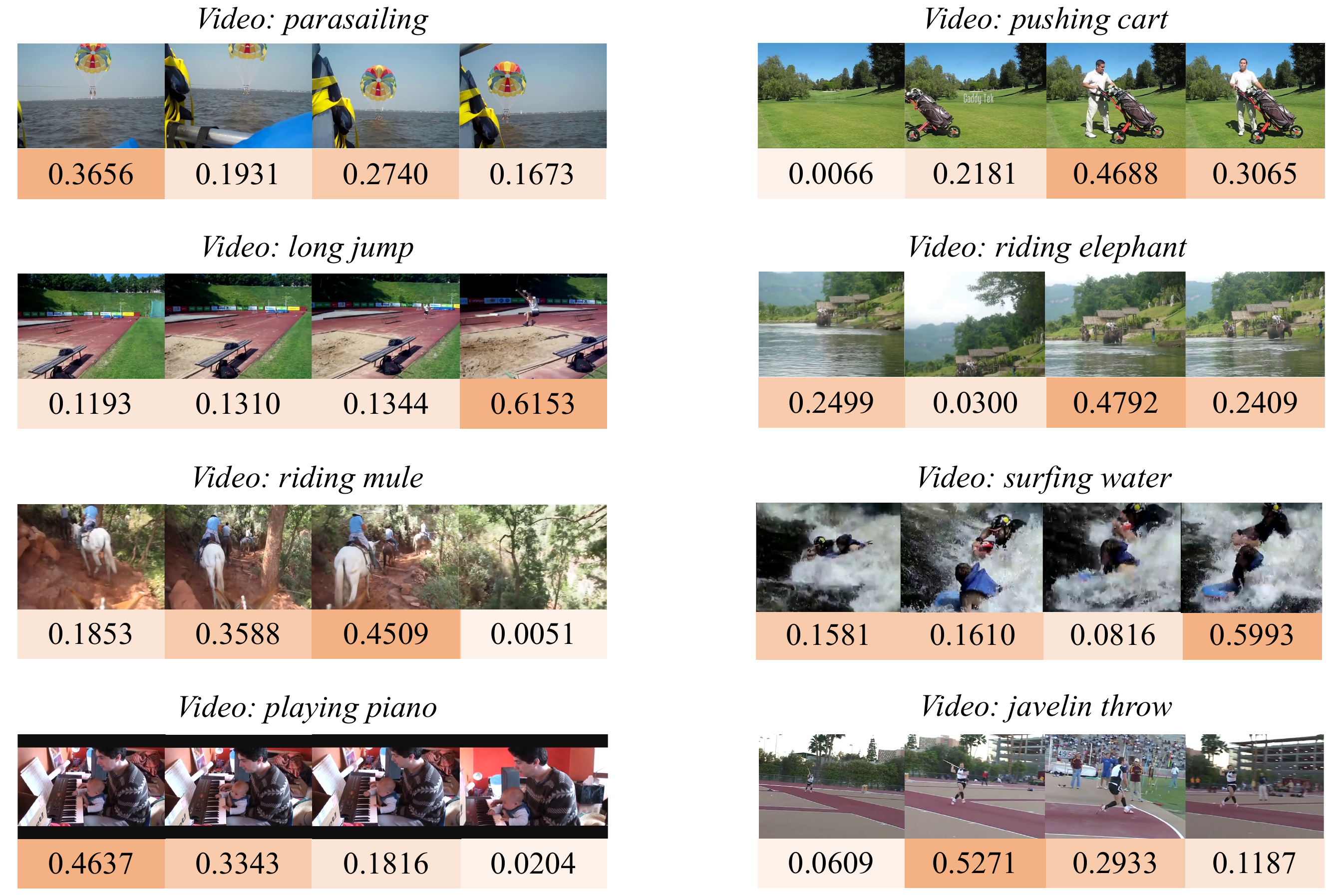}
\end{center}
\caption{Visualization of temporal saliency from our \textbf{Video Concept Spotting} mechanism. Please zoom in for best view.}
\label{fig:sali}
\end{figure*}

\section{More Results}\label{supp:video}

\subsection{Comparisons on UCF-101 and HMDB-51}\label{sec:small_sota}
In this section, we evaluate the performance of our method on the UCF-101 and HMDB-51 datasets to demonstrate its capacity for generalization to smaller datasets.
We finetune our models on these two datasets using the pre-trained ViT-L model on Kinetics-400 and report the accuracy on split one. We use 16 frames as inputs and train for 30 epochs.
Table \ref{t:UCFHMDB} shows that our model has strong transferability, achieving a mean class accuracy of 98.8\% on UCF-101 and 83.1\% on HMDB-51.

\begin{table}[h]
\centering
\begin{tabular}{lcc}
\toprule
\textbf{Method}   & \textbf{UCF-101}  & \textbf{HMDB-51} \\ \midrule

ARTNet~\cite{ARN}  & 94.3\% & 70.9\% \\  
I3D~\cite{i3d}    & 95.6\%  & 74.8\% \\ 
R(2+1)D~\cite{r2+1d}    & 96.8\%  & 74.5\% \\
S3D-G~\cite{s3d} &  96.8\%  & 75.9\% \\ 
TSM~\cite{tsm} &  95.9\%  & 73.5\%  \\
STM~\cite{stm}  &  96.2\%  & 72.2\%   \\ 
MVFNet~\cite{wu2020MVFNet}  &  96.6\%  & 75.7\%  \\ 
TDN~\cite{tdn} & 97.4\% & 76.4\% \\ \midrule
Ours ViT-L  & \textbf{98.8\%} & 82.2\% \\
Ours ViT-L (336$\uparrow$) & 98.6\% & \textbf{83.1\%} \\
\bottomrule
\end{tabular}
\caption{Top-1 accuracy on UCF-101 and HMDB-51 achieved by different methods which are transferred from their \textbf{Kinetics Pre-trained} models with RGB modality.}
\label{t:UCFHMDB}
\end{table}



\begin{figure*}
\centering
\subcaptionbox{Generated attributes from Kinetics-400 lexicon.}{\includegraphics[width = 1\textwidth]{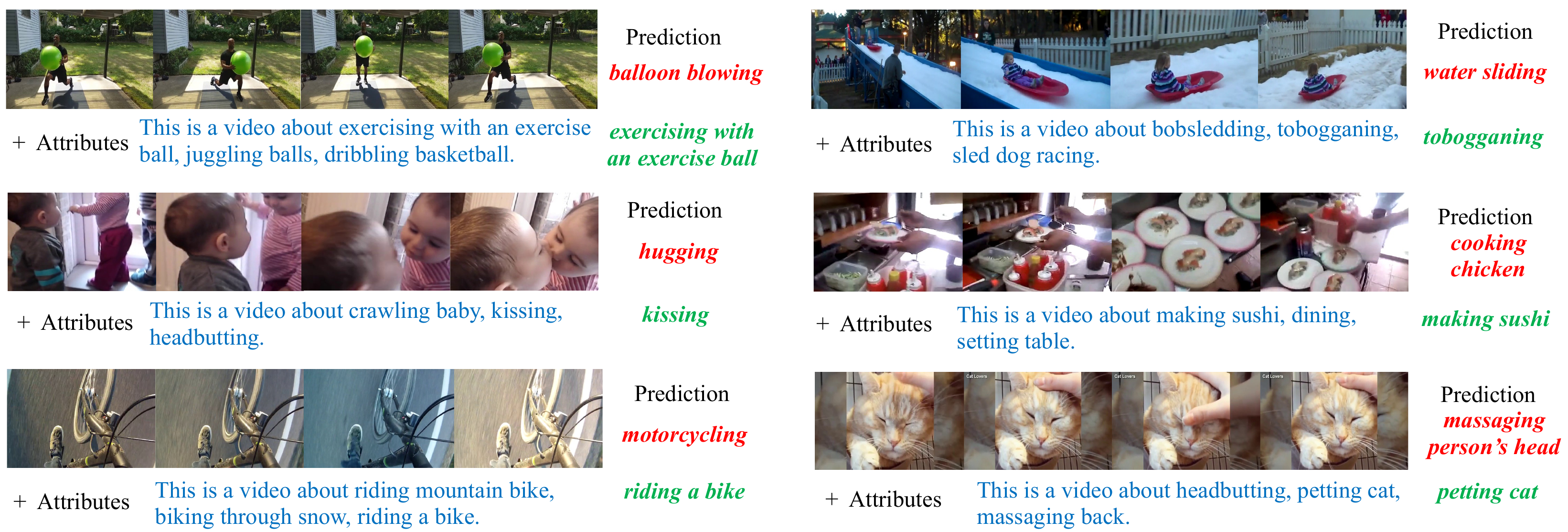}}
\hfill
\subcaptionbox{Generated attributes from ImageNet-1K lexicon.}{\includegraphics[width = 1\textwidth]{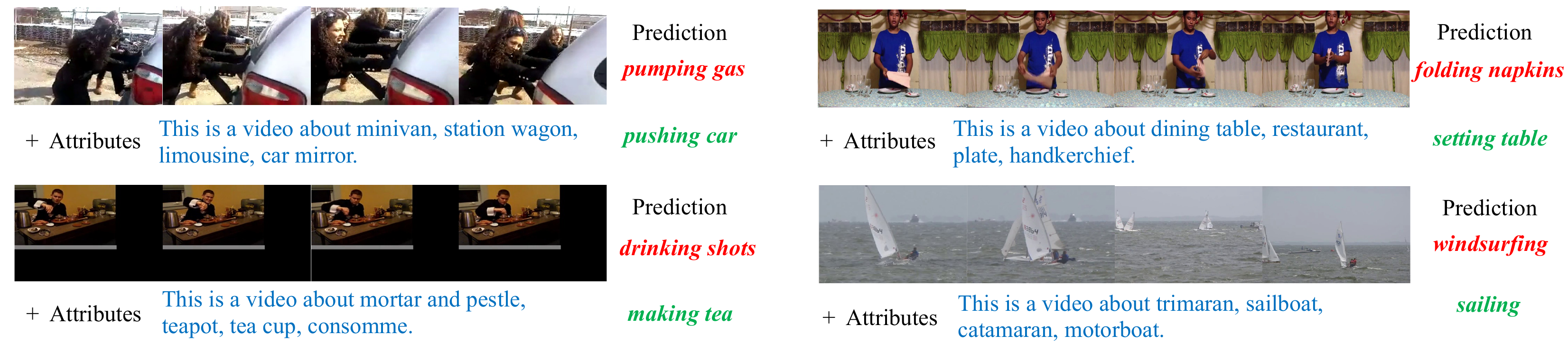}}
\vspace{-0.5em}
\caption{Visualization of the attribute sentence generated by the \textbf{Video-Attribute Association} mechanism that corrected the original {\color{red}incorrect} prediction to the {\color{green}correct} one.}
\label{fig:attr}
\end{figure*}

\subsection{More Qualitative Results}\label{sec:vis}
We present additional visualizations of the \emph{Temporal Saliency} generated by our Video Concept Spotting mechanism in Figure~\ref{fig:sali}. In Figure~\ref{fig:attr}, we also showcase more visualizations of the \emph{Generated Attributes} produced by our Video-Attribute Association mechanism using two different lexicons.

\end{document}